\documentclass[10pt,twocolumn,letterpaper]{article}

\usepackage[pagenumbers]{cvpr} 

\definecolor{cvprblue}{rgb}{0.21,0.49,0.74}
\usepackage[pagebackref,breaklinks,colorlinks,allcolors=cvprblue]{hyperref}
\usepackage{multirow}
\usepackage{graphicx}
\usepackage{caption}

\def\confName{CVPR}
\def\confYear{2027}

\title{Multi-View Relighting with Mergeable Radiance Fields}

\author{
Yinan Wang$^{1,*}$ \quad
Rohan Kadkol$^{2,*}$ \quad
Hyunjin Yoo$^{3}$\\
Jean-Fran\c{c}ois Lalonde$^{2}$ \quad
James J. Clark$^{1}$\\[0.5em]
$^{1}$McGill University, Montreal, QC, Canada\\
$^{2}$Universit\'e Laval, Qu\'ebec, QC, Canada\\
$^{3}$FORVIA, Montreal, QC, Canada\\
{\tt\small yinan.wang2@mail.mcgill.ca \quad
rohan.kadkol.1@ulaval.ca}\\
{\tt\small hyunjin.yoo@forvia.com \quad
jean-francois.lalonde@gel.ulaval.ca}\\
{\tt\small james.clark1@mcgill.ca}\\[0.3em]
{\small $^*$Equal contribution.}
}

\begin{document}   
\maketitle
\begin{figure*}[t]
    \centering
    \includegraphics[width=\textwidth]{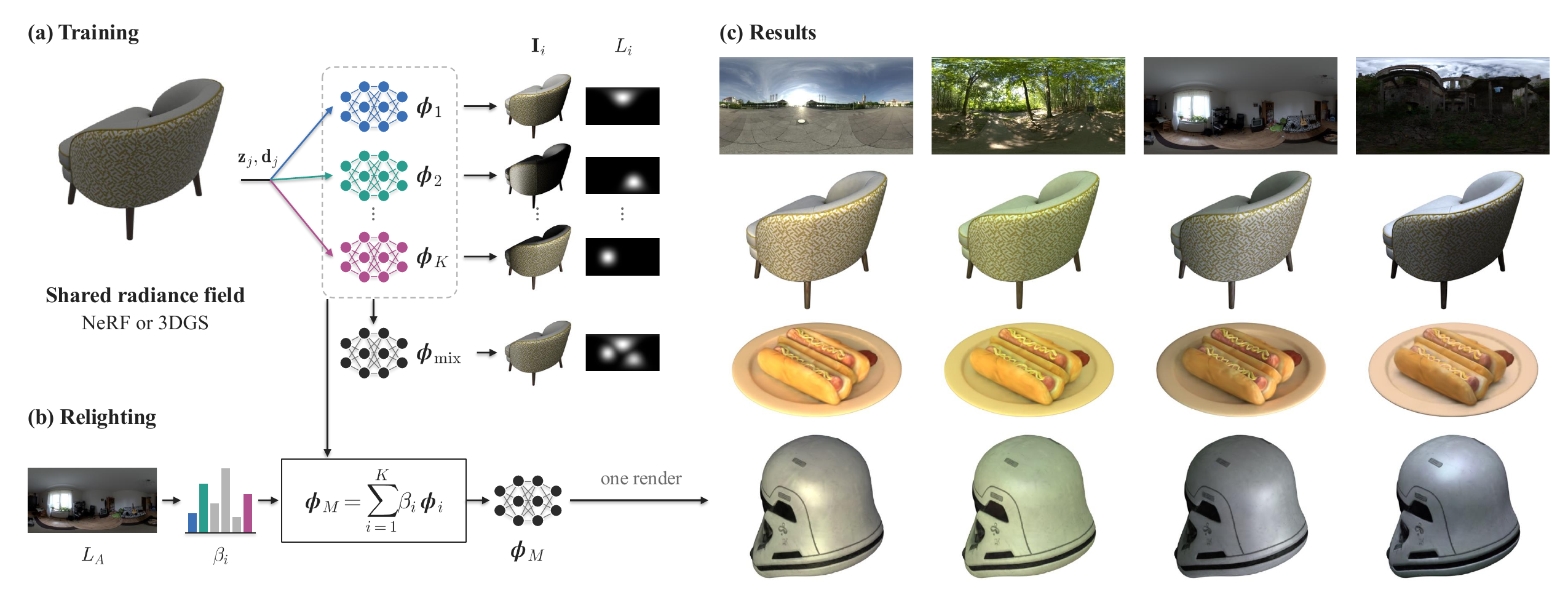}

    \caption{\textbf{Mergeable radiance fields.}
    \textbf{(a)}~One radiance field, a NeRF or a set of 3D Gaussians,
    is shared by $K$ basis illuminations $L_i$. For every element $j$
    that contributes to a pixel (a sample along its ray for NeRF, a
    Gaussian that overlaps it for 3DGS), the field provides a feature
    $\mathbf{z}_j$ and the viewing direction $\mathbf{d}_j$. Each basis
    light has its own small appearance MLP with parameters
    $\boldsymbol{\phi}_i$, trained on the images $\mathbf{I}_i$ of the
    scene under $L_i$. All MLPs start from the same initialization.
    To make them mergeable, some of the training samples are rendered
    with a random convex combination $\boldsymbol{\phi}_{\mathrm{mix}}$
    of the MLPs and supervised by the same combination of the basis
    images, which is the true image under the mixed light (bottom row).
    \textbf{(b)}~To relight, we approximate a new environment map $L_A$
    by a non-negative combination of the basis lights with weights
    $\beta_i$, and take the same weighted sum of the MLP parameters,
    $\boldsymbol{\phi}_M=\sum_i\beta_i\boldsymbol{\phi}_i$.
    The merged MLP $\boldsymbol{\phi}_M$ then renders the relit image
    in a single pass with the unchanged radiance field.
    \textbf{(c)}~Three scenes relit under four unseen environment maps
    at a held-out view, one render each.}

    \label{fig:teaser}
\end{figure*}

\begin{abstract}

We propose MergeableRF, a method for radiance field relighting through model parameter merging. Given multi-view images of a scene under known basis lights, our goal is to relight the scene by directly combining the parameters of radiance fields trained under these lights. This is motivated by the linearity of light transport: the image under a mixture of lights is the same weighted combination of the images under each light. However, combining the parameters of independently trained radiance fields does not produce the corresponding combination of images. To address this, we represent a scene with shared geometry and a separate appearance network for each basis light. We start these networks from the same initialization and introduce merge-aware training, which teaches their parameter combinations to reproduce the corresponding combinations of images. Given a target HDR environment map, we compute its basis-light coefficients and use them to merge the appearance networks into a single relit model. Our method supports both Neural Radiance Fields (NeRFs) and 3D Gaussian Splatting (3DGS), without material estimation, additional relighting networks, or target-specific optimization. Unlike image-space blending, it requires only one rendering pass per viewpoint. Experiments on synthetic Neural Gaffer objects and real OLATverse captures demonstrate relighting quality competitive with methods trained on the same multi-light observations.
\end{abstract}

\section{Introduction}
\label{sec:intro}

Radiance field methods, including Neural Radiance Fields (NeRFs) \cite{mildenhall2021nerf} and 3D Gaussian Splatting (3DGS) \cite{kerbl20233d}, enable realistic novel-view synthesis, but they bake the illumination present at capture time into the appearance of the scene. Changing that illumination, a process known as relighting, therefore requires additional machinery.

Existing relighting methods generally follow two directions. Model-based inverse rendering approaches explicitly estimate scene properties such as geometry, materials, and illumination, and shade them under the new lighting \cite{boss2021nerd,engelhardt2024shinobi,zhang2021nerfactor,jin2023tensoir}. More recent learning-based approaches use generative models or learned appearance representations to synthesize relit images and construct relightable 3D scenes \cite{zhao2024illuminerf,jin2024neural,zhang2025relitlrm,tang2026rogr}. These ideas have also been extended to 3D Gaussian Splatting, where lighting is incorporated through precomputed light transport, explicit shading models, or lighting-conditioned neural representations \cite{zhang2024prtgaussian,li2025recap,fan2025rng,poirier2024diffusion}. Although effective, these methods rely on material and shading models, on representations designed specifically for relighting, or on large generative priors. In this paper, we take a simpler route: we keep the radiance field and its renderer essentially unchanged, and obtain the relit scene by linearly combining network parameters.

Our approach combines two well-known ideas. The first is the linearity of light transport with respect to illumination
\cite{nimeroff1995efficient,debevec2000acquiring}: since light contributions
add, the image under a target illumination can be approximated by a weighted
sum of images under individual basis lights. This holds in \emph{linear
radiance}, \ie when pixel values are proportional to the light reaching the
camera, before any tone mapping, sRGB encoding or clipping.  The second is \emph{model merging} \cite{singh2020model,ainsworth2022git,stoica2024zipit}: under certain conditions, the parameters of neural networks can themselves be combined linearly, and simple weight averaging is effective when the models remain in compatible regions of the loss landscape \cite{wortsman2022model}.

These two forms of linearity do not combine for free. A radiance field is nonlinear in its parameters, so combining the parameters of fields trained under different lights does not, in general, produce the corresponding combination of their images; for independently trained models, we find that it fails entirely. Our key idea is to train parameter combinations of nonlinear appearance networks to reproduce the corresponding combinations of basis-light images. We introduce ``Mergeable Radiance Fields'' (MergeableRF), a framework that trains radiance fields for direct parameter-space merging (\Cref{fig:teaser}).

MergeableRF takes multi-view images of a scene captured under a set of basis lights (\eg, with a light stage) and represents the scene with two components. A single illumination-independent radiance field, either a NeRF or a set of 3D Gaussians, stores the geometry and a feature for every scene point or primitive. One small appearance MLP per basis light decodes this feature and the viewing direction into radiance; for 3DGS, these MLPs replace the conventional spherical harmonics (SH) colors. Training jointly updates the shared geometry and features together with the illumination-specific appearance networks. Since the geometry is shared, only the appearance MLPs need to be merged, and two ingredients make this possible. First, for both NeRF and 3DGS, the basis appearance networks start from identical randomly initialized parameters, which keeps them compatible throughout training \cite{ilharco2022editing,ortiz2023task}. Second, we introduce a merge-aware training strategy: in addition to the usual reconstruction under each basis light, every training step renders random convex combinations of the MLPs and supervises them with the same combinations of the basis-light images. By the linearity of light transport, these are the true images under the mixed lights, so no additional data is needed.

To relight the scene under a target environment map, we express that map as a non-negative combination of the basis lights and combine the appearance MLPs with the same coefficients. Instead of evaluating every basis network and blending the results, MergeableRF renders the merged model directly: once the appearance networks are merged, rendering uses the shared geometry and a fixed number of appearance-network evaluations, independent of the number of basis lights. Changing the target illumination requires no optimization of the scene model.

We evaluate MergeableRF, with both NeRF and 3DGS, under complex high-dynamic-range (HDR) environment maps on synthetic objects from the Neural Gaffer benchmark \cite{jin2024neural} and on real light-stage captures from OLATverse \cite{zhou2026olatverse}, against relighting methods trained on the same basis-lit images. Our analysis separates the error of the lighting basis, of the fitted model, and of merging. With independently initialized networks, the merged render is 14.6\,dB below the blend of the per-light renders; a shared initialization reduces this gap to 8.5\,dB, and merge-aware training to 1.4\,dB. On the synthetic objects, our NeRF model has the second-highest PSNR among the compared methods, and on real captures our 3DGS model is the most accurate.

In summary, our contributions are threefold:
\begin{itemize}
    \item We propose MergeableRF, a relighting framework for NeRF
    and 3DGS that constructs the radiance field for a target
    illumination by linearly combining the parameters of
    basis-light appearance networks over a shared scene
    representation.

    \item We introduce merge-aware supervision, which trains linear
    combinations of appearance-network parameters to reproduce
    the corresponding combinations of basis-light images, at no
    additional data cost.

    \item We demonstrate multi-view relighting under HDR
    illumination on synthetic and real objects with both NeRF
    and 3DGS, and analyze how the lighting basis, the
    initialization, and the mixture supervision each contribute
    to the final error.
\end{itemize}

\section{Related work}
\label{sec:related}

\paragraph{Relighting from basis illuminations.}
For a fixed scene, light transport is linear in illumination. Images under new lighting can therefore be synthesized by blending basis-light images in linear radiance~\cite{nimeroff1995efficient,debevec2000acquiring}. Precomputed radiance transfer (PRT) encodes this relationship in compact bases~\cite{sloan2023precomputed,ng2003all}, while learned transport reduces the observations required for relighting~\cite{xu2018deep,zhang2021neural}. Recent Gaussian methods extend relighting to multi-view light-stage captures~\cite{bi2024gs3,saito2024relightable}.

\paragraph{Relightable NeRFs.}
NeRV~\cite{srinivasan2021nerv}, NeRFactor~\cite{zhang2021nerfactor}, and NeRD~\cite{boss2021nerd} recover scene properties for physically based relighting. TensoIR~\cite{jin2023tensoir} uses tensor factorization to jointly estimate geometry, reflectance, and illumination, with support for shadows and indirect lighting. NeRF-OSR~\cite{rudnev2022nerf} addresses outdoor relighting, while NeRF-W~\cite{martin2021nerf} captures appearance variation without explicit illumination control. Generative approaches, including IllumiNeRF~\cite{zhao2024illuminerf} and Neural Gaffer~\cite{jin2024neural}, use diffusion-based relighting to produce observations for reconstruction under a target illumination.

\paragraph{Relightable 3DGS.}
Gaussian inverse rendering extends these ideas through explicit material and shading models~\cite{liang2024gs,gao2024relightable,jiang2024gaussianshader}. ReCap~\cite{li2025recap} jointly models observations across lighting conditions with shared material properties. SVG-IR~\cite{sun2025svg} introduces spatially varying material and normal attributes within each Gaussian, together with physically based indirect lighting. Other approaches learn light transport more directly. PRTGaussian~\cite{zhang2024prtgaussian} learns per-Gaussian transfer coefficients from multi-view OLAT captures for environment-map relighting. RNG~\cite{fan2025rng} predicts Gaussian radiance conditioned on viewing and lighting directions. OLAT Gaussians~\cite{kuang2024olat} uses learned Gaussian features and neural illumination and scattering functions to model appearance under point lights. MergeableRF places illumination dependence in basis appearance networks and combines their parameters according to the target lighting.

\paragraph{Model merging.}
Parameter averaging can combine networks with compatible weights~\cite{wortsman2022model,frankle2020linear}. Further work addresses parameter alignment, importance weighting, and conflicts during merging~\cite{singh2020model,ainsworth2022git,matena2022merging,yadav2023ties,stoica2024zipit}. Task arithmetic~\cite{ilharco2022editing} and tangent-space analyses~\cite{ortiz2023task} motivate combinations of adaptations around a shared initialization. Closest to our training, Wortsman~\etal~\cite{wortsman2021learning} learn simplexes of networks whose every point is accurate on one task. We also supervise random points of a simplex, but each has its own target: the image under the corresponding mixture of basis lights. These findings support our initialization strategy, but parameter combinations do not generally preserve output combinations. We therefore train merged appearance networks to reproduce the corresponding blends of basis-light images, with scene geometry shared across illuminations.

\paragraph{Relighting datasets.}
Relighting datasets cover human subjects~\cite{nestmeyer2020learning,sun2021nelf,sun2020light,zhou2019deep} and general objects~\cite{ren2015image,xu2018deep}. Object benchmarks include controlled multi-view captures~\cite{bi2020neural,toschi2023relight,liu2023openillumination} and natural environment lighting~\cite{kuang2023stanford}. Neural Gaffer~\cite{jin2024neural} provides synthetic object renderings under diverse HDR environment maps. OLATverse~\cite{zhou2026olatverse} provides multi-view captures of real objects under one-light-at-a-time illumination. We evaluate MergeableRF on these two complementary datasets to assess relighting on synthetic scenes and real captures.

\section{Method}
\label{sec:method}

\subsection{Overview}
\label{sec:overview}
MergeableRF uses a shared scene representation and $K$ appearance networks,
one per basis light $L_i$, whose parameters are linearly combined for
relighting (\Cref{fig:teaser}).
For a fixed scene, light transport is linear in the illumination
when images are represented in linear radiance, where pixel values
are proportional to the light reaching the camera, before tone
mapping, gamma encoding, or clipping~\cite{debevec2000acquiring,
nimeroff1995efficient}. We approximate a
target illumination by a non-negative combination of the basis,
\begin{equation}
L_A \approx \sum_{i=1}^{K}\alpha_i L_i\,,
\qquad \alpha_i \geq 0,
\label{eq:light_combination}
\end{equation}
so that, by linearity, the image under $L_A$ is
\begin{equation}
\mathbf{I}_A \approx \sum_{i=1}^{K}\alpha_i\mathbf{I}_i \,,
\label{eq:image_combination}
\end{equation}
where $\mathbf{I}_i$ is the image under $L_i$. The error of this
approximation is the \emph{basis approximation error}. Evaluating
\Cref{eq:image_combination} directly requires all $K$ appearance networks.
Instead, we combine their parameters, so that relighting needs a single
rendering pass with a merged appearance network (\Cref{sec:inference}). The
discrepancy between this render and the combination of the $K$ per-light
renders is the \emph{merge error}. To keep it small, all appearance networks
start from a shared initialization, and training also supervises merged
networks on sampled mixtures of basis lights (\Cref{sec:training}).

\subsection{Mergeable radiance field representations}
\label{sec:representations}

Both radiance field formulations (NeRF and 3DGS) share an
illumination-independent scene representation, consisting of a geometry
representation and a feature vector $\mathbf{z} \in \mathbb{R}^D$ attached to
every point or primitive. Illumination-specific appearance is represented by a
separate MLP: the color of a scene point with feature $\mathbf{z}$, seen from
direction $\mathbf{d}$ and lit by basis light $L_i$, is
\begin{equation}
\mathbf{c}_i(\mathbf{z},\mathbf{d})
=
f_{\boldsymbol{\phi}_i}\big(\mathbf{z},\gamma(\mathbf{d})\big),
\label{eq:appearance}
\end{equation}
where $\gamma$ is a positional encoding~\cite{mildenhall2021nerf}. All $K$
networks share one architecture: two hidden layers with ReLU activations and a
softplus output, which keeps the predicted HDR radiance non-negative but
unbounded. Network sizes are given in \Cref{sec:exp_setup}.

\paragraph{NeRF.}
Following InstantNGP~\cite{muller2022instant}, the illumination-independent
representation is a multiresolution hash encoding followed by a small MLP,
which outputs a density $\sigma_j$ and a feature vector $\mathbf{z}_j$ at a
scene point $\mathbf{x}_j$. The color of a pixel $\mathbf{p}$ under basis
illumination $L_i$ is
\begin{equation}
\widehat{\mathbf{C}}_i(\mathbf{p})
=
\sum_{j}
w_j f_{\boldsymbol{\phi}_i}(\mathbf{z}_j,\gamma(\mathbf{d}_j)) \,,
\label{eq:nerf_render}
\end{equation}
where the $\mathbf{x}_j$ are sampled on the ray through $\mathbf{p}$,
$\mathbf{d}_j$ is the ray direction, and
$w_j = T_j\big(1-e^{-\sigma_j\delta_j}\big)$ with $T_j$ the accumulated
transmittance~\cite{mildenhall2021nerf}.

\paragraph{3DGS.}
Here, the illumination-independent representation is a set of 3D Gaussians
with the position, covariance and opacity of~\cite{kerbl20233d}; we replace
the spherical-harmonics color of each Gaussian $j$ with a feature
$\mathbf{z}_j$ decoded by \Cref{eq:appearance}. The color of pixel
$\mathbf{p}$ is rendered by alpha compositing, \ie,
\begin{equation}
\widehat{\mathbf{C}}_i(\mathbf{p})
=
\sum_j w_j f_{\boldsymbol{\phi}_i}(\mathbf{z}_j,\gamma(\mathbf{d}_j)) \,,
\label{eq:gs_render}
\end{equation}
where $w_j = a_j\prod_{l<j}(1 - a_l)$, $a_j$ is the opacity of the $j$-th
depth-sorted Gaussian at $\mathbf{p}$, and $\mathbf{d}_j$ is the viewing
direction of that Gaussian.

Although NeRF and 3DGS differ in their scene representations and
compositing weights, both express pixel colors as the same weighted sum of
illumination-dependent MLP outputs. This common formulation allows
us to apply the same parameter-merging strategy to both.

\paragraph{Parameter merging.}
In both formulations, the compositing weights $w_j$ and the features
$\mathbf{z}_j$ do not depend on the illumination. For coefficients
$\beta_i\geq0$ with $\sum_i\beta_i=1$, which describe the contribution of each
basis light to a target illumination, the combination of the $K$ renders is
therefore exactly a single render whose colors are combined per point:
\begin{equation}
\sum_{i=1}^{K}\beta_i\widehat{\mathbf{C}}_i
=
\sum_j w_j\sum_{i=1}^{K}\beta_i\,
f_{\boldsymbol{\phi}_i}\big(\mathbf{z}_j,\gamma(\mathbf{d}_j)\big).
\label{eq:exact_blend}
\end{equation}
We replace the inner sum by a single network with linearly combined
parameters,
\begin{equation}
\boldsymbol{\phi}_M
=
\sum_{i=1}^{K}\beta_i\boldsymbol{\phi}_i,
\label{eq:merge}
\end{equation}
and leave the shared representation unchanged. Because the appearance MLP is
nonlinear in its parameters, the merged network does not generally reproduce
the combination of the basis-light outputs:
\begin{equation}
f_{\boldsymbol{\phi}_M}
\neq
\sum_{i=1}^{K}\beta_i f_{\boldsymbol{\phi}_i}.
\label{eq:merge_error}
\end{equation}
Prior work has shown that models initialized from a common
checkpoint support arithmetic on their
parameters~\cite{ortiz2023task,wortsman2022model,ilharco2022editing}.
We build on this with a shared initialization, and add a merge-aware training
strategy that explicitly supervises merged networks.

\paragraph{Shared initialization.}
Merging independently initialized MLPs produces poor results because their
hidden units are not aligned. We therefore initialize all $K$ appearance MLPs
with identical weights, by copying a single randomly initialized network.
Independently initialized networks reconstruct the individual basis lights
equally well, but their merged outputs are far less accurate
(\Cref{sec:experiments}).

\subsection{Training}
\label{sec:training}

All parameters are trained jointly on linear radiance. Each training
step draws $G$ samples of two kinds, each rendered on a set of training
pixels: a batch of rays for NeRF, one image for 3DGS.

\paragraph{Basis samples.}
A basis sample draws one basis light $i$ at random, renders the model with
$\boldsymbol{\phi}_i$, and compares the result with the training image
$\mathbf{I}_i$ using the standard photometric loss $\mathcal{D}$ of each
representation~\cite{muller2022instant,kerbl20233d}.

\paragraph{Mixture samples.}
To reduce the merge error directly, we also supervise merged networks during
training. A mixture sample draws $k$ basis lights and weights
$\boldsymbol{\beta}$ at random on the simplex. It renders the model with the
merged network
$\boldsymbol{\phi}_{\boldsymbol{\beta}}=\sum_i\beta_i\boldsymbol{\phi}_i$
(\Cref{eq:merge}) and compares it with
$\mathbf{I}_{\boldsymbol{\beta}}=\sum_i\beta_i\mathbf{I}_i$, which is the
exact image under $\sum_i\beta_iL_i$ by linearity, so no additional images are
needed. Gradients reach the $k$ MLPs and the shared representation. Since a
target environment map typically activates many basis lights, we use mixtures
of $k\in\{2,4,8,16\}$ lights.

\paragraph{Objective.}
With $\mathcal{S}$ and $\mathcal{M}$ the basis and mixture samples of a step,
the training objective is
\begin{equation}
\mathcal{L}
=
\frac{1}{G}\Big[
\sum_{g\in\mathcal{S}}
\mathcal{D}\big(\widehat{\mathbf{I}}_{i_g},\mathbf{I}_{i_g}\big)
+
\sum_{g\in\mathcal{M}}
\mathcal{D}\big(\widehat{\mathbf{I}}_{\boldsymbol{\beta}_g},\mathbf{I}_{\boldsymbol{\beta}_g}\big)
\Big].
\label{eq:total_loss}
\end{equation}
We ablate training on basis samples only ($\mathcal{M}=\emptyset$) and the
mixture size $k$ in \Cref{sec:experiments}.

\subsection{Relighting with a target environment map}
\label{sec:inference}

Given an HDR environment map $L_A$, we fit the coefficients independently for
each color channel $c\in\{R,G,B\}$ by non-negative least squares over the
sphere:
\begin{equation}
\boldsymbol{\alpha}_c^{*}
=
\arg\min_{\boldsymbol{\alpha}_c\geq0}
\int_{\mathbb{S}^{2}}
\Big(
L_A^{c}(\boldsymbol{\omega})
-
\sum_{i=1}^{K}\alpha_{i,c}L_i(\boldsymbol{\omega})
\Big)^{2}
d\boldsymbol{\omega},
\label{eq:nnls}
\end{equation}
where $L_i$ carries no channel index because the basis lights are achromatic.
We then separate each channel's intensity from its distribution over the
basis:
\begin{equation}
s_c=\sum_{i=1}^{K}\alpha^{*}_{i,c},
\qquad
\beta_{i,c}=\frac{\alpha^{*}_{i,c}}{s_c}.
\label{eq:channel_normalization}
\end{equation}
Each channel gets its own merged appearance network,
\begin{equation}
\boldsymbol{\phi}_{M,c}
=
\sum_{i=1}^{K}\beta_{i,c}\boldsymbol{\phi}_i,
\label{eq:channel_merge}
\end{equation}
and the relit image is
\begin{equation}
\widehat{\mathbf{C}}_A[c]
=
s_c\sum_j w_j
\Big[f_{\boldsymbol{\phi}_{M,c}}\big(\mathbf{z}_j,\gamma(\mathbf{d}_j)\big)\Big]_c \,.
\label{eq:final_render}
\end{equation}
Because the geometry is shared, \Cref{eq:final_render} needs one ray march
(NeRF) or one rasterization (3DGS), with three appearance-MLP evaluations per
point instead of the $K$ required by \Cref{eq:exact_blend}. Changing the
target illumination requires only a new least-squares fit and a linear
combination of parameters; no optimization of the scene model is needed.

Alternatively, color can be carried by the basis: each basis light is
trained under three color tints, with one appearance network per tinted
light. A single fit over all channels then yields one set of coefficients,
and a single merged network renders all three channels. Our 3DGS model uses
this variant on synthetic scenes (\Cref{sec:exp_setup}).

\subsection{Self-distillation}
\label{sec:distillation}

Mixture training reduces but does not eliminate the merge error. As an
optional refinement, we distill the image-space blend of
\Cref{eq:exact_blend} into the merged appearance
network~\cite{hinton2015distilling}. The trained basis networks provide the
supervision, so no ground-truth image under the target illumination is
required. We initialize the student parameters $\boldsymbol{\phi}$ with the
merged parameters $\boldsymbol{\phi}_M$ and solve
\begin{equation}
\min_{\boldsymbol{\phi}}
\sum_{v\in\mathcal{V}}
\mathcal{D}\Big(
\widehat{\mathbf{I}}_{\boldsymbol{\phi}}^{\,v},
\sum_{i=1}^{K}\beta_i\widehat{\mathbf{I}}_i^{\,v}
\Big),
\label{eq:distill}
\end{equation}
where $\mathcal{V}$ is the set of training viewpoints, and
$\widehat{\mathbf{I}}_{\boldsymbol{\phi}}^{\,v}$ and
$\widehat{\mathbf{I}}_i^{\,v}$ are the images rendered with the student and
with basis network $\boldsymbol{\phi}_i$. Only the student is optimized; the
shared representation and the $K$ basis networks remain fixed. With
per-channel merging, one student is distilled per channel. The distilled
network has the architecture of the merged one, so the rendering cost is
unchanged; unlike direct merging, however, this refinement requires a short
optimization ($3{,}000$ steps) for each target illumination.

\section{Experiments}
\label{sec:experiments}
We evaluate MergeableRF, with NeRF and with 3DGS, on the synthetic objects of
the Neural Gaffer benchmark~\cite{jin2024neural} and on real light-stage
captures from OLATverse~\cite{zhou2026olatverse}, against methods that are also
trained under multiple known illuminations
(\Cref{sec:exp_ng,sec:exp_olat}), and ablate the basis lights and the training
of the appearance networks (\Cref{sec:exp_analysis}).
\subsection{Setup}
\label{sec:exp_setup}

\paragraph{Basis lights.}
\label{sec:basis-lights}
Each basis light is an achromatic von Mises--Fisher lobe
$L_i(\boldsymbol{\omega})=A\exp\!\big(\kappa(\boldsymbol{\omega}^{\top}\boldsymbol{\mu}_i-1)\big)$,
centered on a vertex $\boldsymbol{\mu}_i$ of an icosahedron subdivided zero,
one or two times ($K=12$, $42$, $162$). $\kappa$ follows the spacing of the
lobe centers (half width at half maximum $38^\circ$, $19^\circ$, $9^\circ$),
and $A$ gives all lobes the same integrated radiance. A \emph{broad} 162-lobe
basis keeps the 162 directions with the lobe width of the 42-lobe basis. Our
NeRF model uses $K{=}42$; our 3DGS model uses the broad 162-lobe basis on the
synthetic objects and $K{=}42$ on OLATverse. All images are linear radiance; a
display transform is applied only for evaluation.

\paragraph{Data and evaluation.}
For the six Neural Gaffer objects, we generate Blender renders under every basis light and 101
camera poses (71 for training, 30 held out). We also render the objects under four
target environment maps of the benchmark as ground truth (24 object--map pairs). All training images of an object share one exposure, chosen so that fewer than $10\%$ of its object pixels clip; HDR methods such as ours receive them unclipped, the others receive them clipped and sRGB-encoded. Relit images are compared with the ground truth at the
held-out camera poses after exposure (one per object--map pair, set from the ground truth by the same rule), clipping and sRGB encoding. We report PSNR, SSIM~\cite{wang2004image} and
LPIPS~\cite{zhang2018unreasonable} on object pixels at $256\times256$ (the
centered square of the frame), with the same evaluation for
every method and both datasets.

\paragraph{Models.}
Our NeRF model is a hash-encoded field~\cite{muller2022instant} whose
appearance MLPs have width 64 and read a 16-dimensional feature ($7{,}171$
parameters each). It is trained at $256\times256$ on a centered crop around the object, and its
renders are resampled to the evaluation grid. Each step draws $G{=}14$ batches of 512 rays, 6 under single
basis lights and 8 under mixtures of $k\in\{2,4,8,16\}$ lights, for 315k
iterations (30k on OLATverse); a target map is merged into one network per
color channel (\Cref{eq:channel_merge}). Our 3DGS model follows
\cite{kerbl20233d}, with appearance MLPs of width 128 and a 32-dimensional
feature, trained for 60k iterations with $G{=}24$ images per step, half of them
mixtures of $k\in\{2,4,8,16\}$ lights. It uses each broad lobe with three
color tints (486 basis lights), so a target map is merged into a single
network (\Cref{sec:inference}). On OLATverse it uses the 42 achromatic lobes, merged per color channel, with
appearance MLPs of width 64 and a 16-dimensional feature, trained for 30k
iterations with $G{=}12$ images per step, half of them mixtures of two lights. In both models the mixture samples also update
the shared representation. Details are in the supplement.

\paragraph{Baselines.}
ReCap~\cite{li2025recap}, TensoIR~\cite{jin2023tensoir} and
SVG-IR~\cite{sun2025svg} estimate materials; they receive each basis light as
a known environment map and are relit by shading with the target map.
PRTGaussian~\cite{zhang2024prtgaussian} is relit with the spherical-harmonics
projection of the target map. RNG~\cite{fan2025rng} and OLAT
Gaussians~\cite{kuang2024olat} are point-light methods: they receive each
basis light as a distant light and are relit with one render per basis light,
combined with our fitted weights (\Cref{eq:image_combination}). All are
trained on the same camera views and with the same exposure as our models,
with 12, 42 and 162 basis lights on the synthetic objects (SVG-IR: 12 and 42;
RNG and OLAT Gaussians: 12 only). Each is reported at the basis with its
highest PSNR on the test maps, a choice that favors the baselines; on
OLATverse all use the 42 basis lights.

\subsection{Neural Gaffer Dataset}
\label{sec:exp_ng}
\begin{table}[t]
\centering
\small
\setlength{\tabcolsep}{4pt}
\caption{\textbf{Relighting accuracy on the Neural Gaffer objects}
(6 objects $\times$ 4 unseen target maps $\times$ 30 held-out views).
Every method is trained under known basis lights, and each baseline is
reported at the $K$ with the highest PSNR.
Best, second-best, and third-best distinct values are marked in
\textbf{bold}, \underline{underline}, and \textit{italics}, respectively.
$^\ast$Broad lobes.
${\times}3$: three color tints per lobe (486 basis lights).
$^\ddagger$Evaluated with $K{=}12$ only.}
\label{tab:ng}
\begin{tabular}{@{}lcccc@{}}
\toprule
\textbf{Method} & \textbf{$K$} &
\textbf{PSNR$\uparrow$} &
\textbf{SSIM$\uparrow$} &
\textbf{LPIPS$\downarrow$} \\
\midrule
\multicolumn{5}{@{}l}{\textit{Ours}} \\
MergeableRF, NeRF & 42 & \underline{23.38} & 0.800 & 0.035 \\
MergeableRF, 3DGS & 162$^\ast{\times}3$ & \textit{22.21} & 0.811 & \textit{0.034} \\
\midrule
\multicolumn{5}{@{}l}{\textit{Known basis lights}} \\
TensoIR~\cite{jin2023tensoir} & 12 & 22.20 & 0.772 & 0.040 \\
PRTGaussian~\cite{zhang2024prtgaussian} & 162$^\ast$ & 20.54 & 0.699 & 0.060 \\
SVG-IR~\cite{sun2025svg} & 12 & 21.74 & 0.771 & 0.038 \\
RNG~\cite{fan2025rng} & 12$^\ddagger$ & 20.98 & \textit{0.828} & \underline{0.029} \\
OLAT Gaussians~\cite{kuang2024olat} & 12$^\ddagger$ & 21.49 & \underline{0.857} & \textbf{0.022} \\
ReCap~\cite{li2025recap} & 12 & \textbf{26.62} & \textbf{0.894} & \textbf{0.022} \\
\bottomrule
\end{tabular}
\end{table}

\begin{figure*}[t]
\centering
\setlength{\tabcolsep}{0.6pt}
\renewcommand{\arraystretch}{0.5}
\setlength{\fboxsep}{0pt}
\ifx\ngwlen\undefined\newlength{\ngwlen}\fi
\setlength{\ngwlen}{0.095\textwidth}
\newcommand{\ngw}{\ngwlen}
\newcommand{\ngimg}[2]{\IfFileExists{Fig/experiments/ng2/#1_#2.png}{%
  \includegraphics[width=\ngw]{Fig/experiments/ng2/#1_#2.png}}{%
  \fbox{\begin{minipage}[b][\ngw][c]{\ngw}\centering\tiny pending\end{minipage}}}}
\newcommand{\ngenv}[1]{\begin{minipage}[b][\ngw][c]{\ngw}%
  \includegraphics[width=\ngw]{Fig/experiments/ng2/#1_env.png}\end{minipage}}
\newcommand{\nghead}[1]{{\tiny\shortstack{#1}}}
\newcommand{\ngrow}[1]{%
  \rotatebox{90}{\makebox[\ngw][c]{\scriptsize\textit{#1}}} &
  \ngenv{#1} & \ngimg{#1}{gt} & \ngimg{#1}{ours_nerf_closed} & \ngimg{#1}{ours_3dgs_closed} &
  \ngimg{#1}{recap} & \ngimg{#1}{tensoir_ml} & \ngimg{#1}{svgir} & \ngimg{#1}{olatgs} &
  \ngimg{#1}{rng} & \ngimg{#1}{prtgaussian}}
\begin{tabular}{@{}c@{\hspace{2pt}}cc@{\hspace{3pt}}cc@{\hspace{3pt}}cccccc@{}}
& \nghead{Target\\map} & \nghead{Ground\\truth} & \nghead{Ours\\(NeRF)} & \nghead{Ours\\(3DGS)} &
\nghead{\phantom{g}\\ReCap} & \nghead{\phantom{g}\\TensoIR} & \nghead{\phantom{g}\\SVG-IR} &
\nghead{OLAT\\Gaussians} & \nghead{\phantom{g}\\RNG} & \nghead{PRT-\\Gaussian} \\[1pt]
\ngrow{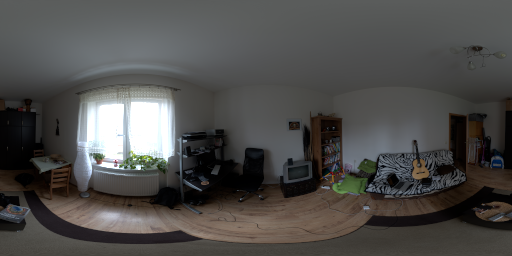} \\
\ngrow{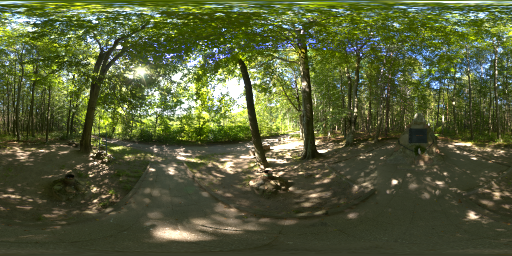} \\
\ngrow{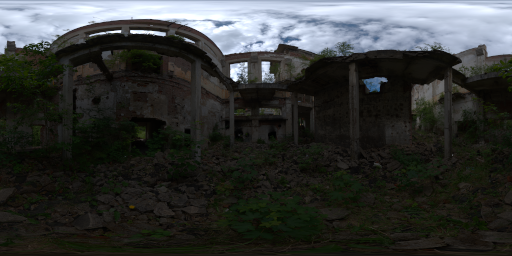} \\
\end{tabular}
\caption{\textbf{Qualitative comparison on the Neural Gaffer objects.} Each
row shows an object relit under a new environment map (left) and seen from a
held-out view. Every method uses the setting reported in \Cref{tab:ng}. Our
results come from a single render of the merged network, with no optimization
for the new lighting. All methods handle the \textit{backpack} well. The shiny
\textit{helmet} and \textit{iron} are harder. Our models get the overall
color and shading right but miss the sharp reflections, because our basis
lights are smooth.}
\label{fig:ng}
\end{figure*}

On the synthetic objects (\Cref{tab:ng} and \Cref{fig:ng}), ReCap, which shades estimated
materials with the target map itself, is the most accurate method. Our NeRF
model has the second-highest PSNR (23.38\,dB), 1.2\,dB above our 3DGS model and
TensoIR (22.21 and 22.20\,dB), which are above SVG-IR, OLAT Gaussians, RNG and
PRTGaussian. Our 3DGS model has a higher SSIM than our NeRF model at a similar
LPIPS. RNG and OLAT Gaussians reach a higher SSIM and a lower LPIPS than both
of our models at a lower PSNR; in our adaptation they are relit with one render
per active basis light, where MergeableRF renders once. For our NeRF model,
merging costs 1.2\,dB relative to the sum of its per-light renders (24.54\,dB)
and raises LPIPS from 0.027 to 0.035.

\subsection{OLATverse Dataset}
\label{sec:exp_olat}

OLATverse~\cite{zhou2026olatverse} contains real objects captured in a light
stage with 331 LEDs. We use six objects of its validation split with different
materials and its benchmark cameras, 24 for training and 6 held out, and
resample the linearized images to $256\times256$ around the object. Since light
transport is linear, the image under an environment map is a weighted sum of
the 331 one-light-at-a-time images, each LED weighted by the integral of the
map over its Voronoi cell. We form the training images under our $K{=}42$
basis lights and the ground truth under five environment maps in this way, so
every method is trained and evaluated on real photographs.

\begin{table}[t]
\centering
\small
\setlength{\tabcolsep}{6pt}
\caption{\textbf{Relighting accuracy on OLATverse} (6 real objects $\times$ 5
environment maps $\times$ 6 held-out cameras). Every method is trained under
the same $K{=}42$ basis lights.}
\label{tab:olat}
\begin{tabular}{@{}lccc@{}}
\toprule
\textbf{Method} & \textbf{PSNR$\uparrow$} & \textbf{SSIM$\uparrow$} & \textbf{LPIPS$\downarrow$} \\
\midrule
\multicolumn{4}{@{}l}{\textit{Ours}} \\
MergeableRF, NeRF & 23.63 & 0.713 & 0.103 \\
MergeableRF, 3DGS & \textbf{24.35} & \textbf{0.836} & \textbf{0.048} \\
\midrule
\multicolumn{4}{@{}l}{\textit{Known basis lights}} \\
PRTGaussian~\cite{zhang2024prtgaussian} & 21.55 & 0.673 & 0.098 \\
RNG~\cite{fan2025rng}                   & 24.32 & 0.807 & 0.069 \\
ReCap~\cite{li2025recap}                & 22.53 & 0.782 & 0.063 \\
\bottomrule
\end{tabular}
\end{table}

On real captures (\Cref{tab:olat}) our 3DGS model matches RNG in PSNR (24.35
\vs 24.32\,dB) with a higher SSIM and a lower LPIPS, and is 1.8\,dB above
ReCap, the most accurate method on the synthetic objects. Our NeRF model is
0.7\,dB below RNG and above ReCap and PRTGaussian in PSNR, with the highest
LPIPS. Merging costs 0.4\,dB for our NeRF model and 1.1\,dB for our 3DGS
model, whereas the sums of their per-light renders are already 9.5 and
8.0\,dB below the sum of the captured basis images: most of the error is
present before the parameters are merged (\Cref{sec:exp_analysis}).

\subsection{Ablation Study}
\label{sec:exp_analysis}

\begin{table}[t]
\centering
\footnotesize
\setlength{\tabcolsep}{3pt}
\caption{\textbf{Ablations} on the Neural Gaffer objects (NeRF, evaluated as in
\Cref{tab:ng}). \emph{Upper bound}: the $K$ ground-truth basis images summed
at the model's training resolution, with no model involved. \emph{Sum of $K$
renders}: the model's $K$ per-light
renders summed. \emph{Merged}: one render of the merged network. $K{=}162$:
broad lobes. $\star$: our configuration (\Cref{sec:exp_setup}).}
\label{tab:ablation}
\resizebox{\columnwidth}{!}{%
\begin{tabular}{@{}lcccccc@{}}
\toprule
& & \textbf{\shortstack{Upper\\bound}} & \textbf{\shortstack{Sum of $K$\\renders}} & \multicolumn{3}{c}{\textbf{Merged (ours)}} \\
\cmidrule(lr){3-3}\cmidrule(lr){4-4}\cmidrule(l){5-7}
& \textbf{Iter.} & \textbf{PSNR$\uparrow$} & \textbf{PSNR$\uparrow$} & \textbf{PSNR$\uparrow$} & \textbf{SSIM$\uparrow$} & \textbf{LPIPS$\downarrow$} \\
\midrule
\multicolumn{7}{@{}l}{\textit{Number of basis lights, pairs ($k{=}2$)}} \\
$K{=}12$                 & 105k & 21.22 & 20.46 & 20.03 & 0.779 & 0.038 \\
$K{=}42$                 & 315k & \textbf{27.96} & \textbf{25.00} & 22.63 & 0.786 & 0.036 \\
$K{=}162$                & 315k & 27.95 & 24.40 & 22.17 & 0.760 & 0.040 \\
\midrule
\multicolumn{7}{@{}l}{\textit{Number of basis lights, $k\in\{2,4,8,16\}$}} \\
$K{=}42$ $\star$         & 315k & \textbf{27.96} & 24.54 & 23.38 & 0.800 & 0.035 \\
$K{=}162$                & 315k & 27.95 & 23.89 & 22.60 & 0.768 & 0.040 \\
$K{=}42$ $\star$ + self-distillation & 315k & \textbf{27.96} & 24.54 & \textbf{24.36} & \textbf{0.836} & \textbf{0.029} \\
\midrule
\multicolumn{7}{@{}l}{\textit{Training of the appearance networks ($K{=}42$)}} \\
Basis only, independent init & 30k & & 24.45 &  9.81 & 0.462 & 0.098 \\
Basis only, shared init      & 30k & & 24.48 & 16.01 & 0.595 & 0.056 \\
Pairs ($k{=}2$)              & 30k & & 24.15 & 21.83 & 0.744 & 0.043 \\
Triplets ($k{=}3$)           & 30k & & 24.12 & 22.46 & 0.765 & 0.040 \\
$k\in\{2,4,8,16\}$ $\star$   & 30k & & 24.16 & 22.81 & 0.779 & 0.038 \\
\bottomrule
\end{tabular}}
\end{table}

\begin{figure}[t]
\centering
\setlength{\tabcolsep}{1pt}
\renewcommand{\arraystretch}{0.6}
\newcommand{\dcimg}[1]{\includegraphics[width=0.232\linewidth]{Fig/experiments/decomp/#1.png}}
\newcommand{\dclab}[1]{\raisebox{0.105\linewidth}{\rotatebox[origin=c]{90}{\tiny #1}}}
\newcommand{\dcrow}[1]{\dclab{NeRF} &
  \dcimg{#1_gt} & \dcimg{#1_basis} & \dcimg{#1_blend} & \dcimg{#1_ours}}
\newcommand{\dcgsrow}[1]{\dclab{3DGS} &
  \dcimg{#1_gt} & \dcimg{#1_gs_basis} & \dcimg{#1_gs_blend} & \dcimg{#1_gs_ours}}
\begin{tabular}{@{}c@{\hspace{1pt}}cccc@{}}
& {\tiny Ground truth} & {\tiny Upper bound} & {\tiny Sum of renders} & {\tiny Merged (ours)} \\
\dcrow{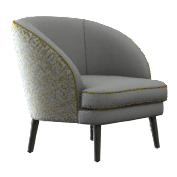} \\
\dcgsrow{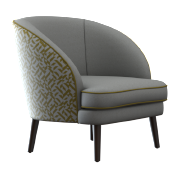} \\
\dcrow{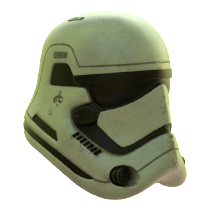} \\
\dcgsrow{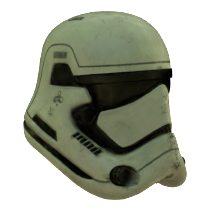} \\
\end{tabular}
\caption{\textbf{Basis error and merge gap} for our NeRF and 3DGS models
(held-out view). \emph{Upper bound}: ground-truth basis images summed with the
fitted weights, with no model involved. \emph{Sum of renders}: the model's
per-light renders summed. \emph{Merged (ours)}: one render of the merged
network.}
\label{fig:decomp}
\end{figure}

\paragraph{Basis error and merge gap.}
\Cref{tab:ablation} separates three stages (\Cref{fig:decomp}).
The \emph{upper bound} combines ground-truth basis images with fitted
weights, measuring basis error without a model. The \emph{sum of $K$
renders} instead combines the model's per-light renders, while
\emph{merged} is our single render. We define the \emph{merge gap}
as the PSNR difference between the last two, measured against ground
truth. This gap does not directly measure their image difference:
the two renders agree at 30.77\,dB PSNR for NeRF and 26.98\,dB for 3DGS.
With 12 lobes, basis error limits quality (upper bound 21.22\,dB).
With 42 lobes, the upper bound improves by 6.7\,dB, but the merge
gap grows from 0.4 to 2.4\,dB as each target map uses 26--40
rather than 2--10 lobes. Increasing to 162 lobes of the same width
does not improve the upper bound and gives each light fewer training
updates, reducing merged PSNR for both pairs and larger mixtures
(22.60 \vs 23.38\,dB). We therefore use 42 lobes for NeRF, while our 3DGS model uses 162 broad lobes with three color tints per lobe (486 basis lights).

\paragraph{Training of the appearance networks.}
With independently initialized MLPs the sum of renders is as accurate as with
a shared initialization (24.45 \vs 24.48\,dB), but the merge collapses
(9.81\,dB). A shared initialization alone still leaves an 8.5\,dB merge gap;
mixtures of two lights cut it to 2.3\,dB. Since a test map combines 26--40
lights, we train with larger mixtures: triplets reduce the gap to 1.7\,dB and
mixtures of 2, 4, 8 and 16 lights to 1.4\,dB, while the sum of renders stays
within 0.4\,dB of basis-only training. At 30k iterations this already exceeds
pairs trained for 315k (22.81 \vs 22.63\,dB); at 315k it reaches 23.38\,dB
with a merge gap of 1.2\,dB.

Self-distillation (\Cref{sec:distillation}) removes most of what remains
(24.36\,dB, against 24.54\,dB for the sum of renders), at the price of a short
optimization for each target map.

\begin{table}[t]
\centering
\footnotesize
\setlength{\tabcolsep}{3pt}
\caption{\textbf{Ablations on the Neural Gaffer objects} (3DGS, evaluated as in
\Cref{tab:ng}; 162 lobes $\times$ 3 tints). \emph{Upper bound}: the
ground-truth basis images summed, here at full resolution, hence higher than
in \Cref{tab:ablation}. \emph{Sum of renders}: the model's 486 per-light
renders summed. \emph{Merged}: one render of the merged network. $\star$: our configuration (\Cref{sec:exp_setup}). Unless stated, mixtures of $k\in\{2,4,8,16\}$ lights.}
\label{tab:gs_ablation}
\resizebox{\columnwidth}{!}{%
\begin{tabular}{@{}lcccccc@{}}
\toprule
& & \textbf{\shortstack{Upper\\bound}} & \textbf{\shortstack{Sum of\\renders}} & \multicolumn{3}{c}{\textbf{Merged (ours)}} \\
\cmidrule(lr){3-3}\cmidrule(lr){4-4}\cmidrule(l){5-7}
& \textbf{Iter.} & \textbf{PSNR$\uparrow$} & \textbf{PSNR$\uparrow$} & \textbf{PSNR$\uparrow$} & \textbf{SSIM$\uparrow$} & \textbf{LPIPS$\downarrow$} \\
\midrule
\multicolumn{7}{@{}l}{\textit{Lobe width (feature 16, width 64)}} \\
Sharp lobes                    & 30k & \textbf{33.60} & 21.84 & 20.34 & 0.763 & 0.041 \\
Broad lobes                    & 30k & 29.07 & 24.06 & 21.97 & 0.799 & 0.037 \\
\midrule
\multicolumn{7}{@{}l}{\textit{Network size and training length (broad lobes)}} \\
Feature 32, width 64           & 30k & 29.07 & 24.20 & 22.01 & 0.800 & 0.038 \\
Feature 32, width 128          & 30k & 29.07 & 24.04 & 22.02 & 0.802 & 0.036 \\
Feature 32, width 128 $\star$  & 60k & 29.07 & 24.68 & 22.21 & 0.811 & 0.034 \\
\midrule
\multicolumn{7}{@{}l}{\textit{Training of the appearance networks (feature 32, width 128)}} \\
No mixture samples             & 60k & 29.07 & \textbf{26.21} & 16.19 & 0.686 & 0.051 \\
$k\in\{2,4,8,16\}$ $\star$     & 60k & 29.07 & 24.68 & 22.21 & 0.811 & 0.034 \\
$\star$ + self-distillation    & 60k & 29.07 & 24.68 & \textbf{24.70} & \textbf{0.862} & \textbf{0.026} \\
\bottomrule
\end{tabular}}
\end{table}

\paragraph{3DGS.}
\Cref{tab:gs_ablation} repeats the analysis for our 3DGS model. Trained
without mixture samples, its sum of renders is 1.5\,dB more accurate than that
of our model, but its merged render reaches only 16.19\,dB; with mixtures of
2, 4, 8 and 16 lights the merge gap is 2.5\,dB, and self-distillation closes
it (24.70\,dB). At the same 162 directions, sharp lobes have a 4.5\,dB higher
upper bound than lobes of the width of the 42-lobe basis, but are harder to
fit: their sum of renders is 2.2\,dB lower and their merged render 1.6\,dB
lower. At 30k iterations a larger network changes neither result
by more than 0.2\,dB; training twice as long adds 0.6\,dB to the sum of
renders and 0.2\,dB to the merged render.

\begin{table}[t]
\centering
\footnotesize
\setlength{\tabcolsep}{3pt}
\caption{\textbf{Ablations on OLATverse} (NeRF, $K{=}42$, 30k iterations; six
real objects, five environment maps, 6 held-out cameras). \emph{Upper bound}:
the captured basis images summed. \emph{Sum of $K$ renders}: the model's 42
per-light renders summed. \emph{Merged}: one render of the merged network.
$\star$: our configuration (\Cref{sec:exp_setup}). The rows change the network size (width, feature size), or the
mixture size.}
\label{tab:olat_ablation}
\resizebox{\columnwidth}{!}{%
\begin{tabular}{@{}lcccc@{}}
\toprule
& \textbf{\shortstack{Sum of $K$\\renders}} & \multicolumn{3}{c}{\textbf{Merged (ours)}} \\
\cmidrule(lr){2-2}\cmidrule(l){3-5}
& \textbf{PSNR$\uparrow$} & \textbf{PSNR$\uparrow$} & \textbf{SSIM$\uparrow$} & \textbf{LPIPS$\downarrow$} \\
\midrule
Upper bound (captured basis images)    & 33.51 & & & \\
\midrule
\multicolumn{5}{@{}l}{\textit{Network size (mixtures $k\in\{2,4,8,16\}$)}} \\
Width 64, feature 16 $\star$           & 24.01 & 23.63 & 0.713 & 0.103 \\
Width 128, feature 16                  & 24.11 & 23.69 & 0.718 & 0.102 \\
Width 64, feature 32                   & 24.23 & 23.82 & 0.721 & 0.101 \\
Width 128, feature 32                  & 24.25 & \textbf{23.85} & 0.720 & 0.101 \\
\midrule
\multicolumn{5}{@{}l}{\textit{Mixture size (width 64, feature 16)}} \\
Pairs ($k{=}2$)                        & \textbf{24.48} & 23.65 & \textbf{0.726} & \textbf{0.098} \\
Triplets ($k{=}3$)                     & 24.27 & \textbf{23.85} & \textbf{0.726} & 0.099 \\
$k\in\{2,4,8,16\}$ $\star$             & 24.01 & 23.63 & 0.713 & 0.103 \\
\bottomrule
\end{tabular}}
\end{table}

\paragraph{Network and mixture size on real captures.}
\Cref{tab:olat_ablation} repeats the analysis on OLATverse. There the merge
gap is small (0.4\,dB), while the sum of renders is 9.5\,dB below the upper
bound (33.5\,dB): the per-light renders at the held-out views, not merging,
limit the result. Larger networks help little: a
32-dimensional feature improves the merged result by 0.2\,dB, width 128 by
0.1\,dB and both together by 0.2\,dB, so we keep the smaller networks. Pairs
fit the individual lights better (sum of renders $+0.5$\,dB) but lose more
when merged (0.8 \vs 0.4\,dB), so they end at the level of $k\in\{2,4,8,16\}$
(23.65 \vs 23.63\,dB); triplets keep the small merge gap and reach 23.85\,dB.
All variants lie within 0.25\,dB of each other, so the benefit of larger
mixtures seen on the synthetic objects does not appear here. See the supplement for per-object results, SSIM/LPIPS values, basis-light figures, training curves, and variants.

\section{Conclusion}
\label{sec:conclusion}
We presented MergeableRF, a method for relighting NeRFs and 3DGS by
directly combining their model parameters. Instead of rendering the
scene separately under multiple lights and combining the resulting
images, our method combines the corresponding appearance networks
into a single model that can render the scene under a new illumination.
To make this possible, we use shared initialization and merge-aware
training to ensure that merging model parameters produces the desired
lighting effects. Experiments on synthetic Neural Gaffer objects and real OLATverse captures show that the merged models closely approximate image-space blending, with PSNR decreases of 1.2--2.5\,dB and 0.4--1.1\,dB, respectively.

\paragraph{Limitations.}
Our method requires multi-view images under known basis lights.
Smooth basis lights limit the reproduction of sharp light sources
and specular reflections. Finer lighting bases could address this
limitation, but combining more networks reduces merging accuracy,
which we leave to future work.

\section{LLM Usage Statement}
Anthropic's Claude was used to assist with code development and debugging, while OpenAI's ChatGPT was used for language editing, grammar correction, reference validation and improving the clarity of the manuscript. All AI-assisted outputs were reviewed and verified by the authors, who take full responsibility for everything presented in this paper.

{
    \small
    \bibliographystyle{ieeenat_fullname}
    \bibliography{main}
}


\end{document}